\documentclass{article}

\usepackage[svgnames]{xcolor}
\usepackage{graphicx}
\usepackage[margin= 0.8in,top=8mm]{geometry}
\usepackage{float}
\usepackage{amsmath}
\usepackage{amssymb}
\usepackage{verbatim}
\usepackage{algorithm}
\usepackage{algorithmic}
\usepackage{epstopdf}

\newcommand{\reals}{\mathbb{R}}

 \providecommand{\scalT}[2]{\left\langle{#1},{#2}\right\rangle}

\newcommand*{\titleAT}{\begingroup
\newlength{\drop}
\drop=0.05\textheight

\includegraphics[scale=1.5]{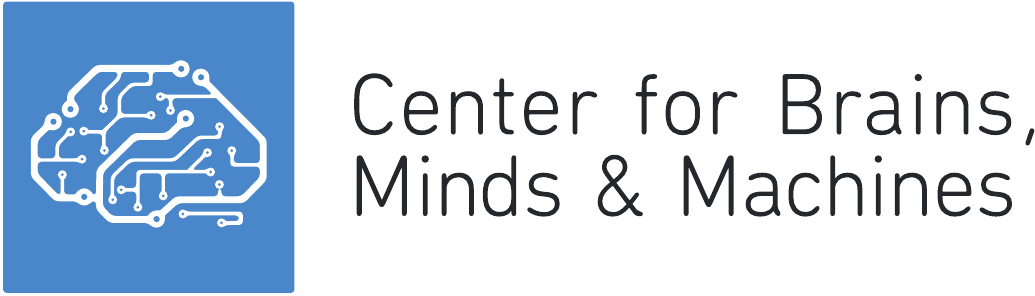}

\textcolor{CornflowerBlue}{\rule{\textwidth}{3 pt}}\par
\vspace{2pt}\vspace{-\baselineskip}
\rule{\textwidth}{0.4pt}\par

\vspace{\drop}
\textbf{\large{CBMM Memo No. \memonumber}}\quad \quad \quad\quad \quad \quad \quad\quad\quad \quad\quad\quad      \textbf{\large{\memodate}}

\vspace{\drop}
\begin{center}
\textbf{\huge{\memotitle}}\\
\vspace{0.4\drop}
\textbf{\Large{by}}\\
\vspace{0.4\drop}
\textbf{\large{\memoauthors}}
\end{center}
\vspace{\drop}
\textbf{\large{\noindent Abstract}:} {\memoabstract}

\textcolor{CornflowerBlue}{\rule{\textwidth}{3 pt}}\par
\vspace{2pt}\vspace{-\baselineskip}
\rule{\textwidth}{0.4pt}\par

\begin{minipage}{.15\linewidth}
\includegraphics[scale=0.1]{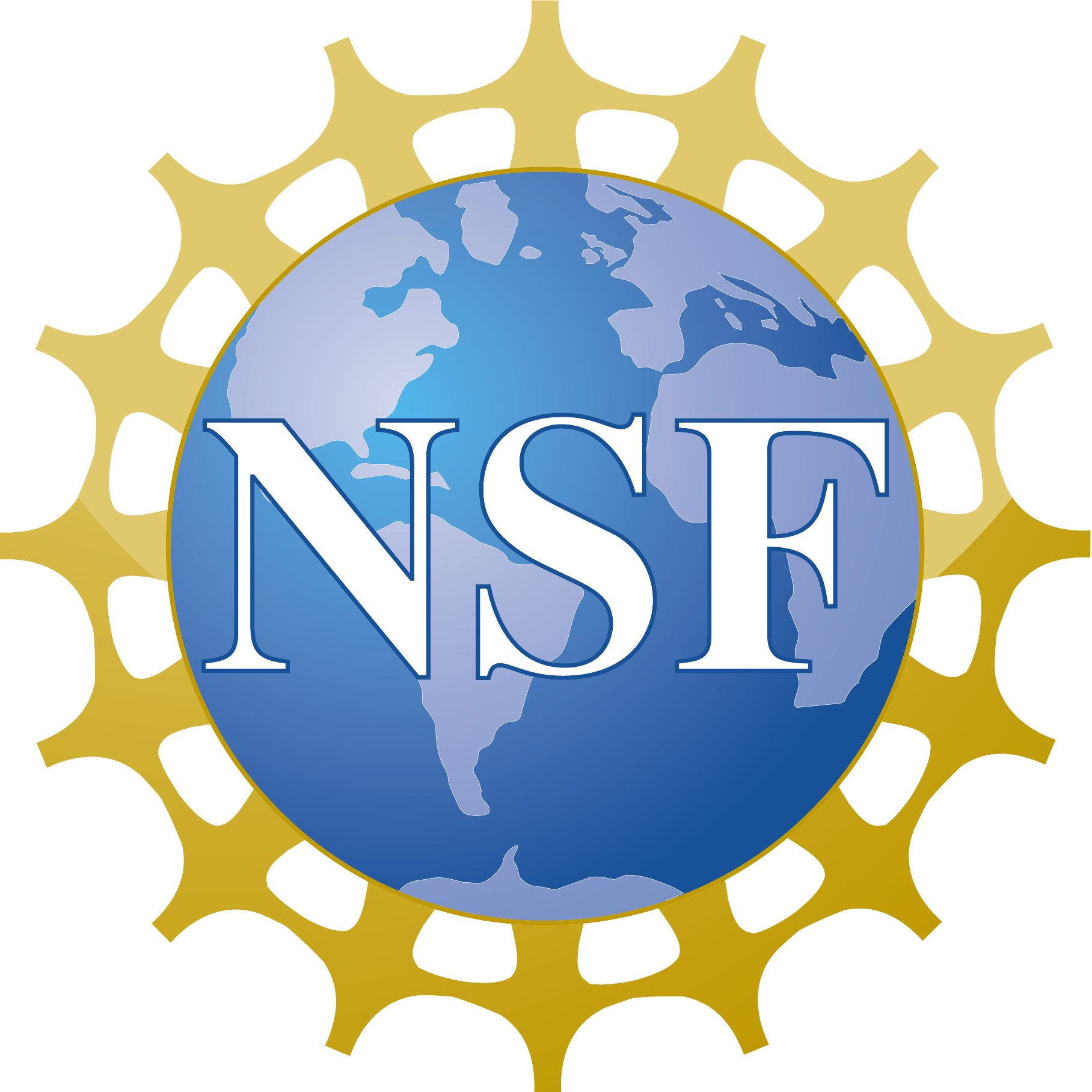}
\end{minipage}
\begin{minipage}{.84\linewidth}
\textbf{\large{This work was supported by the Center for Brains, Minds and Machines (CBMM), funded by NSF STC award  CCF - 1231216.}}
\end{minipage}
\endgroup}

\begin{document}

\pagestyle{empty}

\def\memonumber{23}
\def\memodate{\today}
\def\memotitle{Unsupervised learning of clutter-resistant visual representations from natural videos}
\def\memoauthors{ Qianli Liao, Joel Z Leibo, Tomaso Poggio \\ \text{\normalfont MIT, McGovern Institute, Center for Brains, Minds and Machines}}

\def\memoabstract{Populations of neurons in inferotemporal cortex (IT) maintain an explicit code for object identity that also tolerates transformations of object appearance e.g., position, scale, viewing angle \cite{Hung2005, yamane2008neural, freiwald2010functional}. Though the learning rules are not known, recent results \cite{Li2008,Li2010,li2012neuronal} suggest the operation of an unsupervised temporal-association-based method e.g., Foldiak's trace rule \cite{Foldiak1991}. Such methods exploit the temporal continuity of the visual world by assuming that visual experience over short timescales will tend to have invariant identity content. Thus, by associating representations of frames from nearby times, a representation that tolerates whatever transformations occurred in the video may be achieved. Many previous studies verified that such rules can work in simple situations without background clutter, but the presence of visual clutter has remained problematic for this approach. Here we show that temporal association based on large class-specific filters (templates) avoids the problem of clutter. Our system learns in an unsupervised way from natural videos gathered from the internet, and is able to perform a difficult unconstrained face recognition task on natural images: Labeled Faces in the Wild \cite{Huang2008}.}

\titleAT

\newpage

\section{Introduction}
In natural videos and naturalistic vision the essential properties of images tend to be those that remain stable for some time. Much of what is incidental about images tends to fluctuate more rapidly. Previous efforts to exploit this principle of temporal continuity as a prior assumption enabling unsupervised learning of useful visual representations have only succeeded in demonstrating its effectiveness in rather contrived cases using synthetic objects on uniform backgrounds e.g., \cite{Foldiak1991, wiskott2002slow, Spratling2005, Franzius2011, Isik2012, Rolls2012}. Here we demonstrate a system that is, to the best of our knowledge, the first that can exploit temporal continuity to learn from cluttered natural videos  a visual representation that performs competitively on challenging computer vision benchmarks.

Quite unlike the current ``big data'' trend, which has used systems trained with millions of labeled examples to produce significant advances in object recognition \cite{Krizhevsky2012, Sermanet2013}, our proposal is aimed at understanding one of the strategies used by human visual cortex to learn to see from far less labeled data. To that end, we study how a biologically-plausible feedforward hierarchy can learn useful representations from unlabeled natural videos gathered from the internet. The model we propose is very simple: the only operations it performs are normalized dot products and pooling. It is nonparametric; after an initial layer of Gabor filtering, all the other filters it uses are sampled directly from the unlabeled training data. The classifier we use for the same-different task studied in all but the last section of this article is just a thresholded normalized dot product.

Despite its simplicity, the model performs well on unconstrained face recognition in natural images as well as a basic level task: categorizing lions and tigers. Moreover, in computer vision, solving this face recognition task is generally thought to require a complete detection - alignment - recognition pipeline. Yet the simple model developed here is able to operate directly on the totally unconstrained data: the original unaligned labeled faces in the wild (LFW) dataset \cite{Huang2008}.

\section{Theoretical motivation}\label{section:theory}
Our approach is motivated by a theory of invariance in hierarchical architectures \cite{PartI2013}. As such, it is memory-based and nonparametric. Thus, as proposed by \cite{Isik2012}, the appropriate learning rule to take advantage of the temporal continuity principle is in this case greatly simplified:  just associate temporally adjacent frames. 

Using the notation of \cite{PartI2013}, our system is a Hubel-Wiesel (HW)-architecture. A layer consists of a set of $K$ \emph{HW-modules}. Each HW-module is associated with a \emph{template book} $\mathcal{T}_k$. It is an abstraction of the connectivity that Hubel and Wiesel conjectured gives rise to complex cell receptive fields from simple cell inputs in primary visual cortex \cite{Hubel1962}. From this point of view, one HW-module consists of a single complex cell and all its afferent simple cells\footnote{Note that the simple and complex cells may not actually correspond to real cells, e.g., simple cells might be dendrites and complex cells the soma \cite{PartI2013}. In any case, our usage here is considerably more abstract.}. Each of the $t_k \in \mathcal{T}_k$, called a \emph{template}, is a vector representing an image or a patch of one. 

For an input image $x \in \mathcal{X}$, the ``simple cell operation'' is a normalized dot product \mbox{$\langle x , t \rangle = x \cdot t / \|x\| \| t\|$}. The complex cell ``pools'' the set of simple cell responses with a function $P:\reals^K\rightarrow \reals$ that is invariant to permuting the order of its arguments. The theory of \cite{PartI2013} concerns the case where $P$ is a function that characterizes the distribution of $\langle x, \cdot\rangle$ via its moments, e.g.,  $\max()$ or $\Sigma$. The $k$-th element of the \emph{signature} $\mu : \mathcal{X} \rightarrow \reals^K$ is

\begin{equation}
\mu_k(x) = P\left( \{ \langle x , t \rangle  : t  \in \mathcal{T}_k \} \right).   \label{eq:signature}
\end{equation}
A hierarchical HW-architecture can be constructed by taking the signature of layer $\ell$ as the input to layer $\ell + 1$. 

The temporal association rule we used depends on the choice of a timeframe $\tau$ over which frames will be associated. In this context, we say that two frames are associated when they are in the same template book. Let $X = \{x_1, \hdots, x_{K\tau}\}$ be the full sequence of frames sampled from a training video. Learning is accomplished by initializing HW-modules. Consider a partition of $X$ into $K$ temporally contiguous chunks of duration $\tau$. Set the $k$-th template book $\mathcal{T}_k = \{x_{(k - 1)\tau + 1}, \hdots, x_{k\tau}\}$.

Consider a family of transformations $T$ that has a compact or locally compact group structure. That is, transformations in $T$ are unitary representations of a group $G$. The main theorem of \cite{PartI2013} says that the signature defined by equation \eqref{eq:signature} will be invariant, i.e., $\mu(gx) = \mu(x) ~~ \forall g \in G, x \in \mathcal{X}$, if you choose the template books $\mathcal{T}_k$ to be orbits or fragments of an orbit ($\{gt_k : g \in G\}$) under the action of the group\footnote{Subsets of the orbit are also possible, but the notation becomes more cumbersome so we do not consider that case here. See \cite{PartI2013} for details on this ``partially observed group'' case.}. In this case the actual images in the template books do not matter; they need not resemble the test images. Another theorem from \cite{PartI2013} says that in a more general case where $T$ does not have a group structure, the signature will be approximately invariant. In this case, how close it comes to invariance will depend on the specific templates and their number.

The hypotheses of both theorems can be seen as assumptions about the outcome of the temporal association process. Specifically, they correspond to the case where the training videos show objects undergoing all the transformations in $T$. The theorems give the conditions under which the outcome of that temporal association yields a signature for new images that is invariant to $T$. Simulations based on the theory showed that having the entire orbit of each template is not necessary, all that is needed are samples from it \cite{Liao2013}. 

These considerations imply a potential mechanism for the unsupervised development of invariant visual representations from natural visual experience---or in the case we consider here, from videos gathered from the internet.

\section{Clutter resistance}
\begin{figure*}
\begin{center}

 \includegraphics[width=0.95\linewidth]{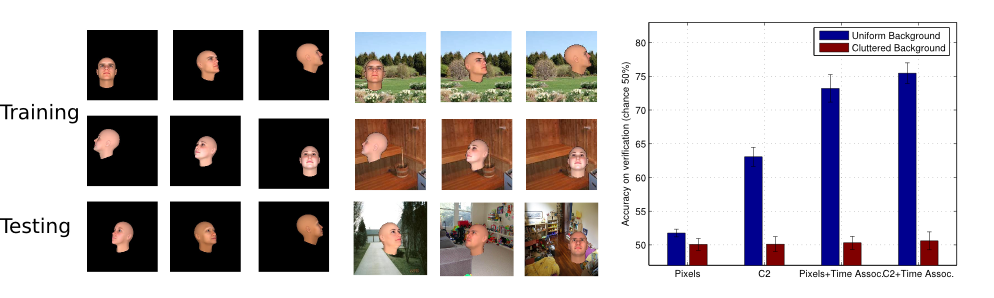} 
\end{center}
   \caption{Temporal association with and without clutter: The upper two rows are examples of the training data. The last row shows the examples of the testing data. The leftmost three columns are example pictures with uniform background. The middle three columns are example pictures with cluttered background. The model is almost the same as that of \cite{Liao2013}, but without doing PCA. The temporal association is modeled by pooling over all the training frames of each individual (e.g., three successive frames on the first row). For the temporal association experiment with clutter, the images of each individual in the training template book have the same background. In the test set, each image has a different background. The C2 features is another type of low-level features, obtained from the second layer of HMAX. The observation is reliable across different low-level features.}
\label{fig:synthetic}
\end{figure*}

Temporal association methods like ours have previously been shown to work in simple uncluttered images \cite{Foldiak1991, wiskott2002slow, Isik2012, Rolls2012}. One may think that this approach to learning invariance is hopelessly stymied by clutter. Figure \ref{fig:synthetic} illustrates the issue. It shows the results from a face pair-matching experiment with synthetic face images from the Subtasks of Unconstrained Face Recognition (SUFR) benchmark \cite{leibo14sufr}: given two images of unknown individuals, the task is to decide if they depict the same person or not. This is the same task used by \cite{Huang2008} though the training procedure differs. The dataset contained 400 faces, with 10,000 images rendered at different orientations (in depth) and different positions in the visual field for each. The reported results were averaged over 10 cross validation folds. Each fold contained 360 training and 40 testing faces.

The training set consisted of 360 videos (transformation sequences)---one for each training individual. Thus, the results in figure \ref{fig:synthetic} were obtained with a one-layer model that had 360 HW-modules; each template book contained all the images of an individual face in the training set.  Note that this uses the label information in the training set to sidestep the important issue of choosing $\tau$. This will not possible in the fully unsupervised case we consider below. 

The classifier was a thresholded normalized dot product (See Section \ref{evaluation_process}). The optimal threshold was determined by maximizing the training accuracy. With this classifier, the raw pixel representation performed at chance. Whereas the pooled representation's performance on the testing set was above 70\% correct. However, when clutter is present, the performance of the temporal association approach drops significantly. This indicates that clutter is indeed a potential problem for these methods. However, in the following sections, we show that it is not insurmountable.

\section{Architectures and simulation methods}
\begin{figure*}
\begin{center}
 \includegraphics[width=1\linewidth]{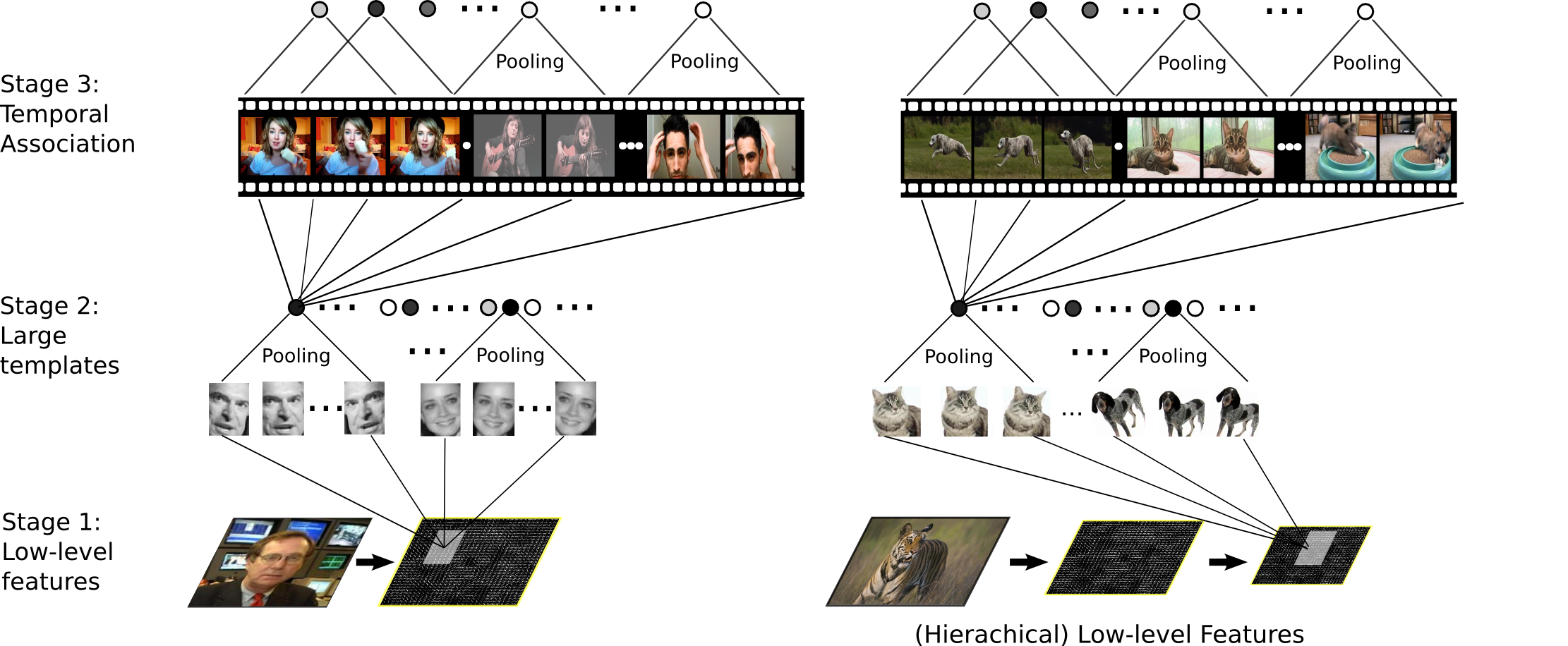}
\end{center}
  \caption{Illustrations of the two models used in this paper. The left one is the face model. It uses closely-cropped faces as the second layer templates. The model on the right hand side is the model for recognizing dogs and cats. It uses closely-cropped cat and dog patches as the second layer templates. The low-level features of the face model is single-layered but that of the latter model is two-layered --- a hierarchical HMAX-like architecture. Empirically, one layer works better for faces since faces are more subtle and require higher resolution.} 
\label{fig:arch}
\end{figure*}

To mitigate the clutter problem, we propose a hierarchical model depicted by Figure \ref{fig:arch}. The architecture consists of a feedforward hierarchy of HW-modules. The final output is the signature $\mu(I)$, a vector of top-level HW-module responses. 

\subsection{Intuition}
The architecture we propose is based on an insight of \cite{Liao2013a}. That work was concerned with replacing the traditional detection-alignment-recognition pipeline used for face recognition with a single end-to-end hierarchy that could identify faces in cluttered natural images like the original \emph{unaligned} labeled faces in the wild data (as opposed to the more commonly used LFWa \cite{Taigman2009} dataset). \cite{Liao2013a} observed that, since most of the distinguishing features of faces are internal (and faces are convex), simple cell responses will rarely be disrupted by the presence of clutter next to the face. Rather, almost all the clutter-related errors could be attributed to ``false positives'', i.e., spuriously high simple cell responses in the background. To see why, consider a small template of just a few pixels, say 2 $\times$ 2. Such a small template will be activated by many non-faces, more importantly: the problem becomes less and less severe as its size is increased. They found that, at least in the case of faces, a layer with a wide spatial pooling range and high-resolution, highly selective, class-specific templates was effective at mitigating this clutter problem. This layer acts like a gate that prevents non-faces from influencing the representation provided to the next layer. 

The face recognition model used in the present work consisted of three layers of HW-modules (see figure \ref{fig:arch}-left). The first ``low-level features'' layer used Gabor templates in a manner similar to the C1 layer of HMAX \cite{Riesenhuber1999, Serre2007a}. The second layer used large class-specific templates with a wide pooling range in space, scale, and in-plane rotation. The final layer was trained by the temporal association method described in section \ref{section:theory} using natural video data gathered from YouTube.

\subsection{Biological motivation}
We do not explicitly model the temporal association process in layers 1 and 2. However, we do not regard this work as a departure from the idea of a unified architecture with similar learning rules operating at each stage. Layer 2's pooling domains: space, scale, and in-plane rotation could be learned by temporal association of videos of faces undergoing those transformations. The layer 1 Gabor templates can also be obtained by temporal association. This is one of the main results of the spectral version of \cite{PartI2013}'s theory. That theory considers more biologically-plausible HW-modules that use principal components of the template books since they arise naturally as a consequence of normalized Hebbian plasticity \cite{oja1982simplified}. See \cite{Anselmi2013a} for details of the spectral theory.

The face-specific part of the system resembles the macaque face-recognition hierarchy \cite{freiwald2010functional}. Like neurons in the middle funds and middle lateral face patches, If one were to record from cells in the model's second layer (the first face-specific layer), they would find them to be tuned to specific face views. The model's final layer resembles the anterior medial face patch in that it can be used to decode face identity despite changes in viewpoint (3D depth rotation)\footnote{In the macaque face recognition hierarchy there is also an intermediate stage between the view-tuned and view-tolerant representations. Interestingly, the intermediate representation in the anterior lateral patch is mirror symmetric. That is, its cells respond nearly identically to pairs of face images flipped over their vertical midline \cite{freiwald2010functional}. This intriguing result may be explained by the spectral theory in \cite{Anselmi2013a} and \cite{Leibo2013a}-chapter5.}.

Most of the motivation for this model came from face recognition studies. However, it is a general-purpose object recognition algorithm. Nothing prevents us from investigating its performance on non-face, basic-level categorization tasks. In section \ref{subsection:basiclevel_results} we report results on the task of distinguishing lions from tigers using a model trained with natural videos of cats and dogs.

\subsection{Summary of architecture details}

The model is trained in a layer-wise fashion. For training, (1) compute layer 1 features of all video frames and large image patches mentioned above. The latter become layer 2 templates. (2) compute layer 2 features of all video frames. Use these for the layer 3 templates.\\

Example patches are shown in the second layers of Figure \ref{fig:arch}. They were generated by sampling large image patches on a face/cats\&dogs dataset. Their low-level feature representations were used as the second layer templates.

Details of both architectures are given in the supplementary information.

\subsection{Training videos and data}
\begin{figure*}
\begin{center}
 \includegraphics[width=0.97\linewidth]{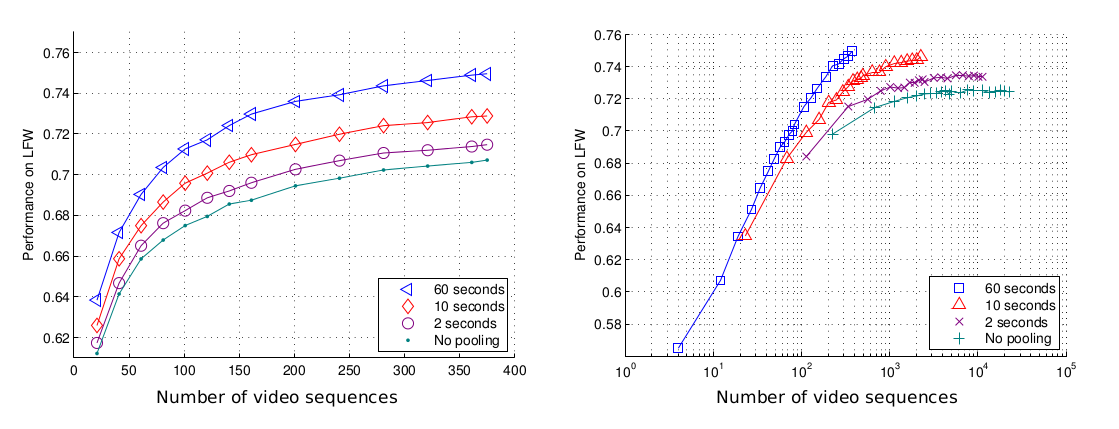} 
\end{center}
\caption{Performance vs. temporal pooling range. There are 375 face videos. In this particular experiment, each video is truncated such that they are all 60 seconds long. The videos are sampled at 1 frame per second. Thus, for each video we get 60 frames. There are 4 pooling options: 60 seconds, 10 seconds, 2 seconds, no pooling. The pooling ranges here are non-overlapping to maximize data usage (but generally overlapping in other experiments of the paper). We observe that the longer the pooling range, the higher the performance is.  We show two different views of the result in this figure -- one is short range and the other is long-range (and log scale). Note that truncation to 60 seconds is only performed in this experiment but not in the final model, and it is only for speed purpose. The pattern is robust across different settings.} 
\label{fig:LFW_main}
\end{figure*}

 In our experiments, the face model learns from 375 videos of faces from YouTube. The lion\&tiger model learns from 224 videos of dogs and cats from YouTube. These videos may contain things other than faces, cats and dogs. See figure \ref{fig:gallery}.\\
\textbf{Simple cells}: Each frame corresponds to a simple cell (in the third layer of our model). For speed purpose, videos are sampled at some slow rates: 0.5, 1 or 2, 4, etc. frames/second. So there are 0.5, 1 or 2, 4, etc. simple cells on average per second.\\
\textbf{Complex cells}: How are the complex cells placed over the time domain? Each complex cell has a pooling domain over time that may or may not overlap with other complex cells. The placement of complex cells over the time domain is a hyper parameter that depends on the experiments. We have two ways of placing the complex cells in this paper:

\begin{enumerate}
\item For the experiment in Figure \ref{fig:LFW_main}, we truncate each video to 60 seconds and place the complex cells in an even and non-overlapping way. This is mainly for speed purposes since none of the simple cells are wasted in this case. See the caption of Figure \ref{fig:LFW_main} for details.
\item For the final performance in Table \ref{table:final_performance}, in order to avoid being biased by some very long videos, we simply specify that each video has exactly $M$ complex cells. Their pooling domains are equally large over time, and they are spread out evenly in each video to maximize the data (i.e., simple cell) usage. For some short videos, there may be some overlaps between the pooling domains of its complex cells. For some long videos, some frames/simple cells are wasted.
\end{enumerate}

\begin{table*}
  \begin{center}
  \begin{tabular}{|cc|cc||p{3.3cm}c|}
 \hline  
\multicolumn{6}{|c|}{ LFW Results } \\
 \hline  
\multicolumn{4}{|c||}{Detected \& Cropped} &  \multicolumn{2}{c|}{Undetected \& Uncropped} \\
 \hline  
   \multicolumn{2}{|c|}{Unsupervised} &  \multicolumn{2}{c||}{Supervised \& Unaligned} &  \multicolumn{2}{c|}{Unaligned} \\
 \hline  
{\bf \scriptsize Model}  & {\bf Acc.}  & {\bf \scriptsize Model}  & {\bf Acc.} & {\bf \tiny Model }  & {\bf \tiny Acc.}      \\
 {\scriptsize  LHS (aligned) \cite{sharma:ECCV2012} }& {\scriptsize 73.4}\%     & {\scriptsize MERL \cite{huang2008lfw} }& {\scriptsize  70.52}\% & {\tiny SIFT-BoW+SVM \cite{Liao2013a} } &      {\scriptsize 57.73$\pm$2.53}\% \\
 {\scriptsize  MRF-MLBP (aligned) \cite{Arashloo2013MRFs} }& {\scriptsize   80.08\%}   & {\scriptsize Nowak et al. \cite{nowak2007learning} } & {\scriptsize 72.45}\%     & {\tiny Our Model(Gabor) }& {\scriptsize 75.57$\pm$1.63}\%\\
 {\scriptsize  I-LPQ (aligned) \cite{Hussain2012}} & {\scriptsize 86.2}\%    &  {\scriptsize   Sanderson et al. \cite{sanderson2009multi} }&  {\scriptsize 72.95}\% & {\tiny Our Model (fusion)} & {\scriptsize  \textbf{81.32$\pm$1.40}\%}\\ 
 {\scriptsize  PAF (aligned) \cite{yi2013towards} } & {\scriptsize 87.77}\%   & {\scriptsize  APEM (fusion) \cite{cuifusing} }& {\scriptsize 81.70}\% & {\tiny Our Model (fusion)+SVM} & {\scriptsize  \textbf{83.23$\pm$1.07}\%} \\
 \hline  
\end{tabular}
\vskip 0.1in
\caption{Note that all top unsupervised methods on LFW require detected, cropped and aligned faces. The SVM results were obtained by replacing the cosine classifier with an SVM. In the final experiment (fusion) in Table \ref{table:final_performance}, we trained three pipelines based on Gabor, PCA and HOG features. The signatures of the third layer of each pipeline were computed separately. They were then fused by a weighted concatenation and fed into the classifier. The concatenation weights were determined by minimizing the training error. }
\label{table:final_performance}
\end{center}
\end{table*}

\subsection{Evaluation Process}
\label{evaluation_process}
Here we briefly describe how the model is evaluated. 
\\
\textbf{Face verification}
Given a pair of images $(x_a,x_b)$ the task is to verify whether they depict the same person or not.  To test our HW-architecture, we run it on both images and compare them by the angle between their signatures (top-level representations).  That is, we take the normalized dot product $\scalT{\mu(x_a)}{\mu(x_b)}$ if it exceeds a threshold $\tau$, our method outputs that $x_a$ and $x_b$ have the same identity otherwise they are different. We use the training set to pick the optimal $\tau$.
\\
\textbf{Lion and tiger classification}
Given an image $x$, the task is to determine whether it depicts a lion or tiger. In this case, we trained an SVM on the final output of our feature extraction pipeline.

\section{Experiments}

\subsection{Subordinate level identification of faces}\label{subsection:face_results}

The model was trained with 375 natural videos gathered from YouTube and tested on the \emph{unaligned} Labeled Faces in the Wild dataset. \cite{Liao2013a} also considered this problem with an approach similar to ours. However, their system was considerably more complicated, using locality-sensitive hashing, and much less biological. Moreover, it was not trained with the unsupervised procedure that is our focus here.

\begin{figure*}[t]
\begin{center}
 \includegraphics[width=.85\linewidth]{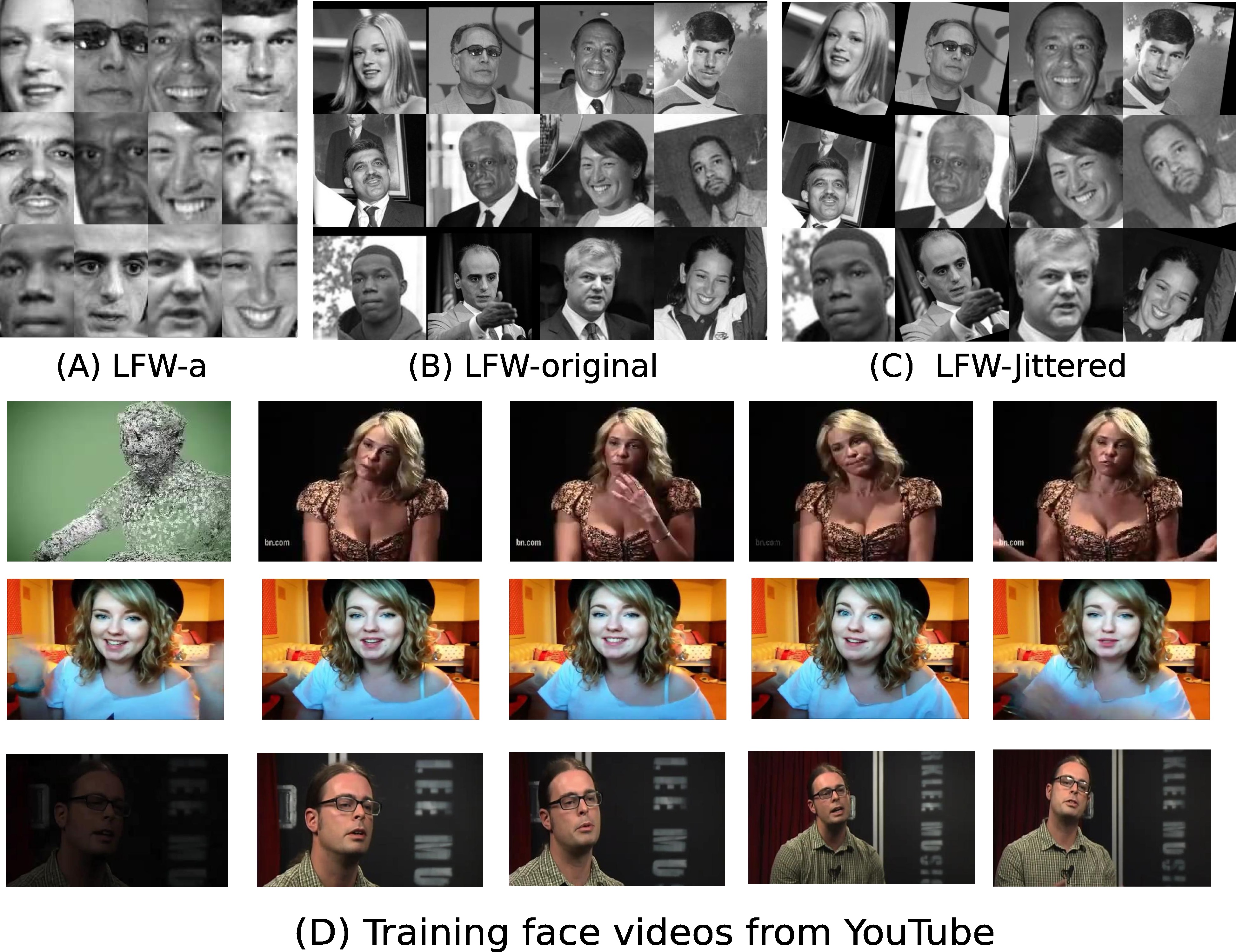} 
\caption{Example testing and training images: (A) LFW-a (LFW-aligned) is the usual dataset used by LFW practitioners. Here we address (B) and (C), which are much more difficult. (D) are some example frames from our training videos. } 
\label{fig:gallery}
\end{center}
\end{figure*}

Example training and testing images are shown in Figure \ref{fig:gallery}. Figure \ref{fig:LFW_main} shows the model's performance for different choices of the temporal association time window $\tau$. It shows that longer temporal association windows perform better. This result confirms the main hypothesis with which this paper is concerned: a hierarchy including a layer of high resolution class-specific templates can be used to mitigate the clutter problem enough so as to enable the learning of class-specific invariances in a subsequent layer.

We also tested the face recognition model on a more difficult dataset---LFW-Jittered \cite{Liao2013a} to assess the model's tolerance to affine transformations (table \ref{table:jittered_performance}). Table \ref{table:jittered_performance} compares our model to others in the literature. Despite its simple unsupervised learning mechanism, our model performs comparably with the best published results on the unaligned dataset. Note however, the unaligned LFW dataset is much less popular than its aligned version. It is unlikely that our model is really as close to the current state of the art as it appears in this table.

\begin{table*}
 \begin{center}
  \begin{tabular}{|c|cc|}
 \hline  
{\bf Model}  & {\bf LFW}  & {\bf LFW-J} \\
 \hline  
 HOG+SVM \cite{Liao2013a}  & 74.45$\pm$1.79/67.32$\pm$1.59\%     & 55.28$\pm$2.02\% \\
 Our Model (Gabor)   & 75.57$\pm$1.63\%  &  75.48$\pm$1.60\% \\
 Our Model (Fusion)  & 81.32$\pm$1.40\%  &  81.10$\pm$1.15\% \\ 
  \hline  
\end{tabular}

\caption{We report the performance of our model on the LFW-J (jittered) dataset created by \cite{Liao2013a}. The LFW-J dataset is created by randomly translating, scaling and rotating the original LFW images. The HOG baseline is from \cite{Liao2013a}.  $74.45\%$ is the closely cropped performance and $67.32\%$ is the non-cropped performance.}
\label{table:jittered_performance}
\end{center}
\end{table*}

\begin{figure*}
\begin{center}
 \includegraphics[width=.98\linewidth]{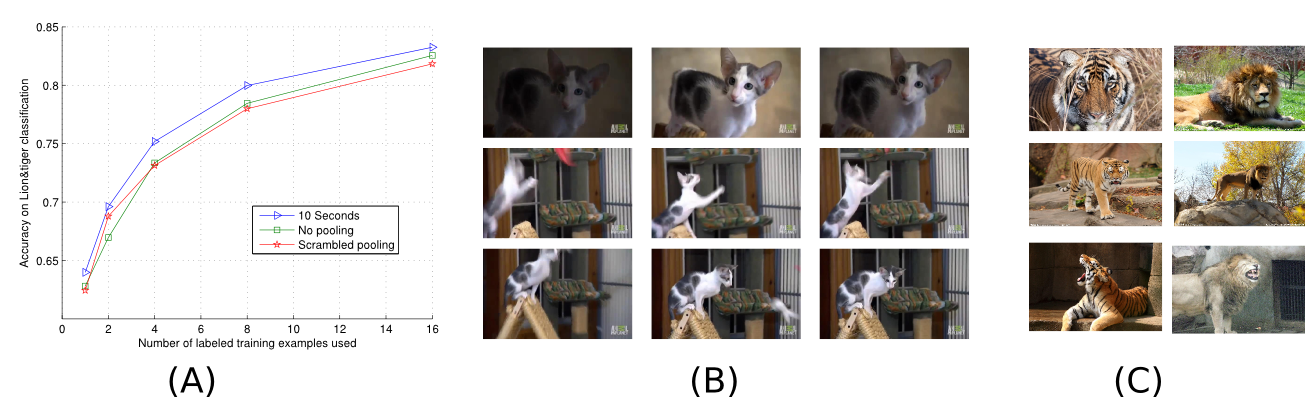} 
\end{center}
\caption{This model was trained using 224 natural videos of cats and dogs from YouTube. In this experiment, frames were sampled at 1 fps. (A) The performance of our model. We tested 10-second pooling, no pooling and 10-second pooling with scrambled frames. The performance is averaged over 120 trials. (B) example training frames from the YouTube videos we gathered. (C) example images from the lion \& tiger testing dataset. } 
\label{fig:lion}
\end{figure*}

\subsection{Basic level categorization: lions and tigers}\label{subsection:basiclevel_results}
The goal of temporal association learning is to produce an invariant representation for the image transformations that naturally arise as objects move relative to their observer. So far we have concentrated on a subordinate level task: face recognition. Such tasks which require very similar objects to be discriminated from one another are thought to depend more critically on invariance. Specifically, they are thought to require more complicated class-specific invariances that would be hard to learn in any way other than from natural videos, e.g., rotation in depth or the transformation of a frown to a smile. In contrast, basic level categorization may rely more on discriminative features that appear similarly from any angle than on sophisticated invariance to class-specific transformations. It follows that temporal association should be a more effective strategy in the subordinate level case. 

To further explore this direction, we applied our model on a basic level categorization task --- categorizing lions and tigers. Figure \ref{fig:lion}) gives the results. We developed our model with 224 natural videos of cats and dogs from YouTube. We also created a lion\&tiger dataset consisting of 1128 images. For training, only 16 labeled examples were used from the lion\&tiger dataset. 400 testing images were used per category.  For the second layer templates, we used large patches from dogs and cats (as shown in \ref{fig:arch}). We found that when pooling over scrambled frames, the performance drops to the level of ``no pooling''. The results are averaged over 120 trials and the effect is robust to different choices of parameters.

\section{Discussion}
One goal to which this work has been directed is that of building a model of the macaque face recognition system that could be used to motivate new experiments. This approach to neuroscience research is based on the idea that neurobiological models of parts of the brain that are thought to perform computational functions (like object recognition) ought to be able to actually perform these functions. Furthermore, the same type of training data available to an infant's developing visual system should be used. In the case of invariant object / face recognition, the exact amount of ``label''-esque information that is available is debatable. But it is surely less than the amount of labeled data used to train state of the art systems like \cite{Krizhevsky2012}. We regard the main contribution of the present work to be its establishing that---at least for faces---it is possible to learn using only minimal supervision, a representation that is robust enough to work in the real world.

\newpage

\appendix \textbf{Supplementary Information}

\section{Architecture details}

Example feature dimensionalities are given as $y_i \times x_i \times z_i$, where $y$ denotes the height, $x$ denotes the width, $z$ denotes the thickness/feature size, and $i$ is the layer number. Our model is architecturally akin to HMAX and Convolutional Neural Networks. All of these architectures can be described using this notation.

Layer 1 features are computed by applying a function $L()$ to the input image (``L'' means low-level features). The function $L()$ transforms the input image ($y_0$ $\times$ $x_0$ $\times$ 3, i.e., colored image) to ($y_1$ $\times$ $x_1$ $\times$ $z_1$), where $y_1$ and $x_1$ are usually smaller than $y_0$ and $x_0$, respectively, while $z_1$ is almost always larger than 3.

We sampled videos of length $T$ seconds at a rate of $f$ frames/second to get $T*f$ frames. Let $F_i$ denote a single frame, where $i \in \{1,\hdots,T*f$\}.

\subsection{Model for face recognition}
\textbf{Training Phase}
\begin{enumerate}
\item \textbf{Preparation:} We randomly chose 5500 images from the SUFR-W \cite{leibo14sufr} dataset. From each of them, we sampled a large image patch (the central 104x80 pixels). For each patch, we generated its -18, -9, 0, 9, 18 degree in-plane rotations and a horizontal flip for each. Hence, we get $5500*10=55000$ large image patches in total, which we collectively call $P$. We use $P_j$ to denote a single patch, with $j \in \{1,\hdots,55000\}$.

We resized each video frame $F_i$ to a height of 400 pixels (with aspect ratio preserved). We then built a multi-scale pyramid by resizing it to 20 scales with resizing ratios 0.26, 0.28, 0.32, 0.36, ..., 1.0. We used $F_i^{pyr}$ to denote the multi-scale pyramid of frame $F_i$.

\item \textbf{Low-level features:} In this experiment, we tried three types of low-level features $L()$: (1) one layer Gabor filtering + 2x2 spatial max pooling with step size 2, (2) HOG features with cell size 8, and (3), one layer of PCA (i.e. projecting to eigenvectors of natural image patches) + 2x2 spatial max pooling with step size 2. All of these features were extracted from gray-scale images.

\item \textbf{First Layer:} We applied the low-level feature function $L()$ to patches $P$ and multi-scale image pyramid $F_i^{pyr}$ to get $L(P)$ and $L(F_i^{pyr})$, respectively, where $L(F_i^{pyr})$ is a pyramid of 20 scales of low-level features. An example pyramid would be of size $x_1^1$ $\times$ $y_1^1$ $\times$ $z_1, ..., x_1^{20}$ $\times$ $y_1^{20}$ $\times$ $z_1$, where the superscripts denote the scales. $L(P)$ then become the second layer templates (i.e., ``simple cells'').

\item \textbf{Second Layer:} ``Simple Cell'' Convolution: each scale of the pyramid $L(F_i^{pyr})$ is convolved with the second layer templates $L(P)$ separately. This is similar to a convolutional layer in a convolutional neural network (CNN) like \cite{Krizhevsky2012} with three differences: (1) we have multiple scales; (2) our templates (i.e., filters) are very large, e..g.,13  $\times$ 10  $\times$  $z_{i-1}$, while that of typical CNNs are very small (e.g., 3 $\times$ 3 $\times$ $z_{i-1}$ or 5  $\times$ 5  $\times$ $z_{i-1}$), where $z_{i-1}$ denotes the thickness of the previous layer; and (3) we use a normalized dot product while CNNs use a dot product followed by a rectifying nonlinearity. Note that the normalized dot product is also nonlinear.

``Complex cell'' pooling: resulting from the above convolution is a pyramid of size $x_2^1 $ $\times$ $ y_2^1 $ $\times$ $ z_2, ..., x_2^s $ $\times$ $ y_2^s $ $\times$ $ z_2$, where $s=20$. For each template, there was a pyramid of size $x_2^1 $ $\times$ $ y_2^1 $ $\times$ $ t, ..., x_2^s $ $\times$ $ y_2^s $ $\times$ $ t$, where in this case $s=20$, and $t=10$. $t$ is the number of in-plane transformations (e.g., rotations and flips) generated in the preparation stage. For each template, we pooled over all scales, locations and in-plane transformations with $max()$ to get a scalar. The output of the second layer was the concatenation of these results. In this case, it was of size 5500.

\item \textbf{Third Layer:} For each video frame, we performed the following steps: 1. encode it in the model up to the second layer. 2. store the layer-2 encodings as a template (``simple cell'') for the 3rd layer.

The face model learned from 375 videos of faces from YouTube. The sampling rate was 1 fps for Figure 3 and 0.5 fps for Table 1. The performance was robust to different sample rates using the same temporal pooling range.

\end{enumerate} 
\textbf{Testing Phase}
\begin{enumerate}
\item \textbf{First Layer:} For a test image, we compute its low-level features as described above.
\item \textbf{Second Layer:} The output of the first layer is convolved with $L(P)$. Then we pool over scales, locations and in-plane transformations as described above. 

\item \textbf{Third Layer:} For each output from the second layer, compute the normalized dot product with the stored third layer training templates (i.e., ``simple cells''). Then pool the responses over temporally adjacent frames over $N$ seconds, where $N$ is a hyper parameter. Let the sampling rate be $f$ frames/second, then there are $N*f$ ``simple cells'' connected to a ``complex cell''. A normalized dot product is performed between the input and the ``simple cells'', generating a 1-D vector of responses.  The ``complex cell'' pools over this vector to get a scalar (using mean pooling) or a vector (using other pooling methods). The final output of the third layer is the concatenation of the responses for all ``complex cells''.

From one perspective, this method is a generalization of local pooling from the spatial domain to the time domain. Each ``complex cell'' has a pooling domain over time that may or may not overlap with other ``complex cells''. 
  
\end{enumerate}

\subsection{Model for dog/cat recognition}
In the following section, we detail our model for recognizing dogs and cats.

\textbf{Training Phase}
\begin{enumerate}
\item \textbf{Preparation:} We prepared 7879 images of cats and dogs and randomly sampled a large image patch (104x104 pixels) from each of them. For each patch, we generated -18, -9, 0, 9, 18 degree in-plane rotations. Hence, we get $7879*5=39395$ large image patches in total, which we collectively call $P$. We use $P_j$ to denote a single patch, where $j \in \{1,\hdots,39395\}$.

We resized each video frame $F_i$ to a image of height 231 pixels (with aspect ratio preserved). We then buillt a multi-scale pyramid by resizing it to 3 scales with resizing ratios 0.8, 0.9 and 1.0. Let $F_i^{pyr}$ denote the multi-scale pyramid of frame $F_i$.

\item \textbf{Low-level features:} In this experiment, $L()$ was a two-layer HMAX-like architecture: (1) the first layer extracted a concatenation of Gabor and PCA features similar to those of the face model, but applied on color images. (2) The output of the first layer was convolved with 4000 templates using a normalized dot product. The templates were $5 \times 5 \times 192$ patches sampled randomly from the first layer features of random natural images gathered from internet, where 192 was the thickness/feature size of the first layer. (3) There is a 2x2 max pooling stage with step size 2. Next, dimensionality reduction by PCA was performed over the template dimension. Note that here the term ``layer'' in this paragraph only refers to the internal architectures of $L()$. We have only tried one type of $L()$ in our experiments, and these modeling choices and parameters are mostly arbitrary and not fine-tuned for performance. Further studies could explore the effects of using other features such as CNNs.

\item \textbf{First Layer:} We applied the low-level feature function $L()$ to patches $P$ and the multi-scale image pyramid $F_i^{pyr}$ to get $L(P)$ and $L(F_i^{pyr})$, respectively, where $L(F_i^{pyr})$ is a pyramid of 3 scales of low-level features. An example pyramid would be of size $x_1^1$ $\times$ $y_1^1$ $\times$ $z_1, ..., x_1^s$ $\times$ $y_1^s$ $\times$ $z_1$, where $s=3$. $L(P)$ was used as the second layer templates.

\item \textbf{Second Layer:}``Simple Cell'' convolution: each scale of the pyramid $L(F_i^{pyr})$ was convolved with the second layer templates $L(P)$ separately.``Complex cell'' Pooling: the result of the above convolution is a pyramid of size $x_2^1 $ $\times$ $ y_2^1 $ $\times$ $ z_2, ..., x_2^s $ $\times$ $ y_2^s $ $\times$ $ z_2$, where $s=3$. For each template, there was a pyramid of size $x_2^1 $ $\times$ $ y_2^1 $ $\times$ $ t, ..., x_2^s $ $\times$ $ y_2^s $ $\times$ $ t$, where $s=3$, and $t=5$. $t$ is the number of in-plane rotations generated in the preparation stage. For each template, we pooled over all the scales, locations, and (in-plane) rotation angles with $max()$ to get a scalar. The output of the second layer is of size 7879.

\item \textbf{Third Layer:} The same procedure was used as for the face model. This model learned from 224 videos of dogs and cats from YouTube. Frames were sampled at 1 fps. 

\end{enumerate} 
\textbf{Testing Phase}
Same as for the face model.

\subsection{PCA approximation}
Note that the matrix $P$, consisting of all the second layer templates (as column or row vectors), is a low rank matrix. We can perform PCA on $P$ and keep the $k$ largest eigenvectors. By projecting the templates and the windows to the $k$ dimensional space defined by those eigenvectors we can perform faster dot products in the reduced dimensional space. This procedure is adopted by \cite{Liao2013a} and similar to the one employed by \cite{chikkerur2011approximations}. It may be biologically plausible in the sense of the spectral theory of \cite{Anselmi2013a}.

\section{Face selectivity of the second layer}
\begin{figure}
\begin{center}
 \includegraphics[width=.8\linewidth]{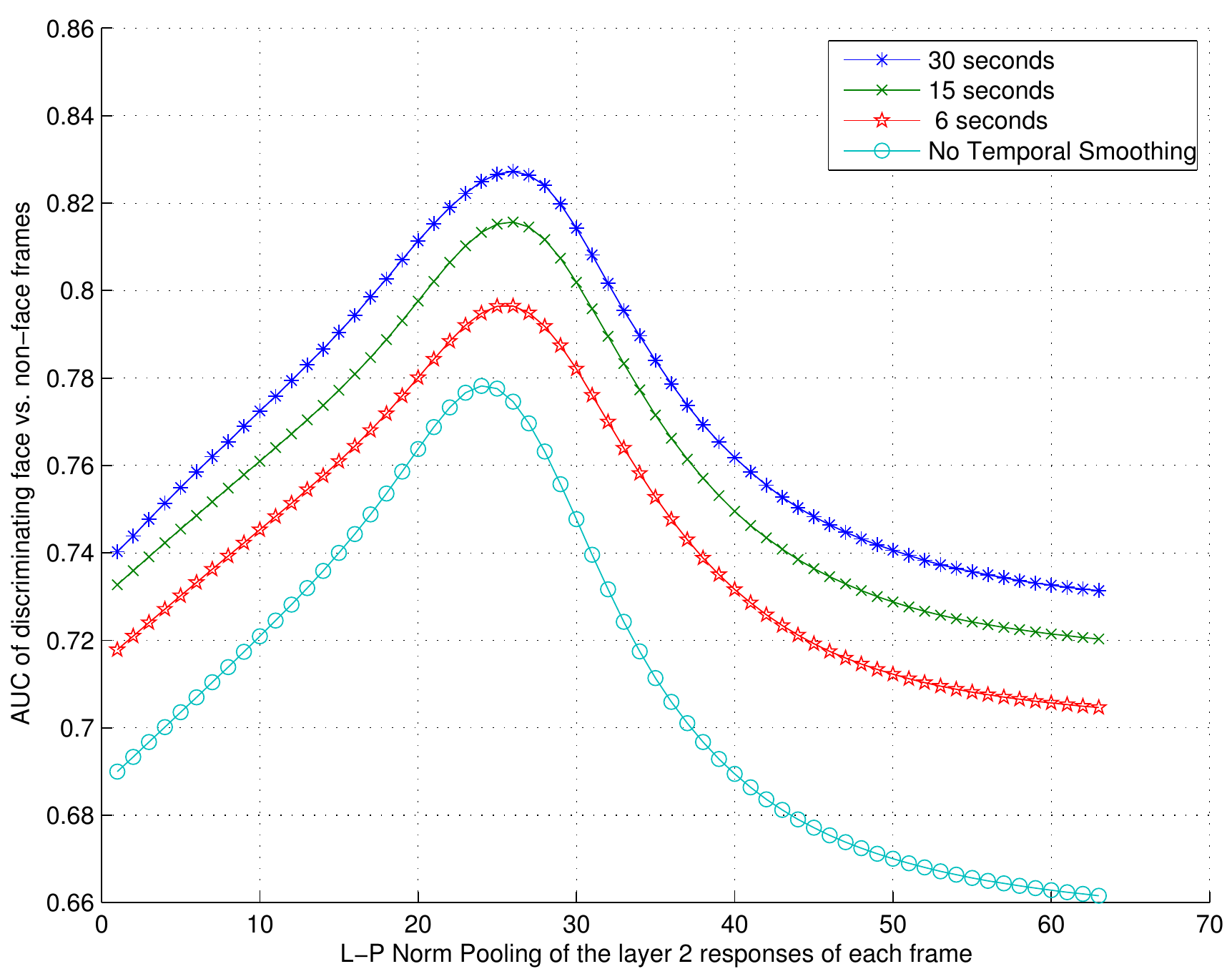} 
\end{center}
\caption{We ran the face model on 22413 frames from the face dataset and 22691 frames from the cat-dog dataset. For each frame, we used the L-P norm to pool over all layer 2 template responses, and thus get one score per frame. When $P = 1$, the pooling was equivalent to mean (sum) pooling. When P approaches infinity, the pooling was equivalent to max pooling (for non-negative numbers). For evaluation purpose, we assumed every frame from the face dataset contains a face, and that none of the frames from the cat-dog dataset contain a face. We computed the ROC curves with this information and the scores obtained above. The AUCs are shown as a function of the order P of the L-P norm.  One explanation for the curve is that for the task of detection, mean and max are not good measures of a match. Somewhere between mean and max is (apparently) better.} 
\label{fig:gating}
\end{figure}

We tested our model's selectivity to face versus non-face frames with a simple mechanism---pooling over layer 2 responses (figure \ref{fig:gating}).

\small{
\bibliographystyle{ieeetr}
\bibliography{bibtex}
}

\end{document}